\documentclass[11pt]{article}

\usepackage[preprint]{acl}

\usepackage{times}
\usepackage{latexsym}
\usepackage{enumitem}
\usepackage{amsmath}
\usepackage{bbm}

\usepackage[T1]{fontenc}

\usepackage[utf8]{inputenc}

\usepackage{microtype}

\usepackage{inconsolata}

\usepackage{graphicx}
\usepackage{listings}
\usepackage{booktabs}

\newcommand{\benchabbr}{DETOUR}
\newcommand{\bench}{\textbf{D}ual-agent based \textbf{E}valuation \textbf{T}hrough \textbf{O}bscure \textbf{U}nder-specified \textbf{R}etrieval}

\title{\benchabbr: An Interactive Benchmark for Dual-Agent\\Search and Reasoning}

\author{
 \textbf{Li Siyan\textsuperscript{1,2}},
 \textbf{Darshan Deshpande \textsuperscript{1}},
 \textbf{Anand Kannappan\textsuperscript{1}},
 \textbf{Rebecca Qian\textsuperscript{1}}
\\
\textsuperscript{1} Patronus AI, \textsuperscript{2} DAP Lab, Columbia University
}

\usepackage[dvipsnames]{xcolor}

\usepackage{todonotes}

\begin{document}
\maketitle
\begin{abstract}

When recalling information in conversation, people often arrive at the recollection after multiple turns. However, existing benchmarks for evaluating agent capabilities in such tip-of-the-tongue search processes are restricted to single-turn settings. To more realistically simulate tip-of-the-tongue search, we introduce \bench~(\benchabbr), a dual-agent evaluation benchmark containing 1,011 prompts. The benchmark design involves a \textit{Primary Agent}, which is the subject of evaluation, tasked with identifying the recollected entity through querying a \textit{Memory Agent} that is held consistent across evaluations. Our results indicate that current state-of-the-art models still struggle with our benchmark, only achieving 36\% accuracy when evaluated on all modalities (text, image, audio, and video), highlighting the importance of enhancing capabilities in underspecified scenarios. 


\end{abstract}

\section{Introduction}

Large Language Model (LLM) assisted search has recently found applications in several domains, including scientific research~\cite{phan2025humanity, lin2025scirgen}, finance~\cite{li2024agent}, customer service~\cite{lyu2025deepshop}, and conversational context collection~\cite{miroyan2025search}. This popularity has given rise to deep research agentic systems~\cite{xu2025comprehensive} that actively acquire and critically reason over content on the internet with the aim to satisfy user queries. Several benchmarks have attempted to study the search capabilities of such agents under conflicting knowledge~\cite{cattan2025dragged} and multimodal~\cite{mialon2023gaia} scenarios, emphasizing the gap between expectations and current model performance. One such important domain is tip-of-tongue (ToT) search~\cite{sky2025browsing,arguello2021tip} that involves web search over vague user recollections of entity descriptions. Most benchmarks in the search evaluation space focus solely on single-turn questions; however, in realistic user conversations, structured ambiguity clarifications for ToT tasks are crucial. A skill that existing benchmarks do not test for is the ability to ask high-quality clarification questions in underspecified scenarios. 

\begin{figure}[!t]
    \centering
    \includegraphics[width=0.9\linewidth]{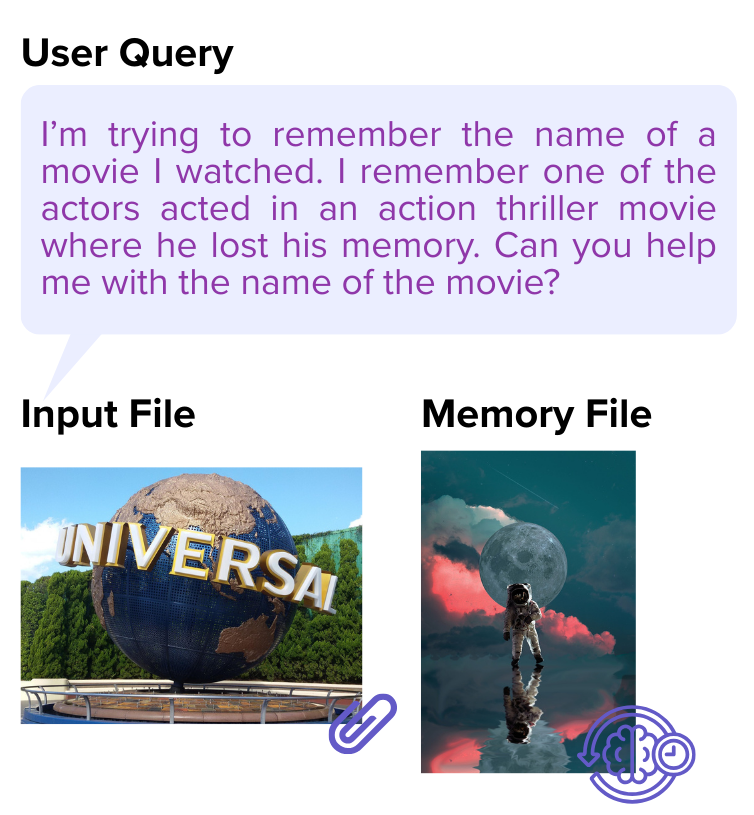}
    \caption{An example instance from \benchabbr. Here, the input file shows the Universal Studios Globe, signaling a \textit{Universal} release. The Memory File includes an astronaut, hinting that the film is set in space. Tying the amnesia action-thriller clue to \textit{The Bourne Identity} leads to \textit{Matt Damon}, and then to \textbf{The Martian}, where his character fights to survive on Mars.}
    \label{fig:ex_bench}
\end{figure}

To address this gap, we propose \bench~(\benchabbr), a collection of \textbf{1,011} complex and diverse queries spanning \textbf{eight domains}, with input modalities including \textbf{image, audio, and video}. \benchabbr~features a frontier evaluation setup that accounts for realistic query ambiguities and incentivizes clarification questions by simulating user memory via a secondary agent. Building on~\citet{sky2025browsing}, we ground the evaluation in memory files. A \textit{Memory Agent} accesses these files containing additional information about the user's recollection without exposing the ground truth to the evaluated agent. This setup helps map the dynamics of a multi-turn, multi-modal search evaluation as opposed to the previously proposed single-turn static tests.

We study the following research questions in detail in this paper:
\begin{enumerate}[itemsep=0.1cm]
    \item How effectively can LLM agents identify and resolve ambiguities through follow-up clarifications in our interactive, tip-of-the-tongue search setting?
    \item Can LLM agents effectively utilize the Memory Agent to achieve better task performance?
    \item What are failure cases emerging from the multi-turn evaluation regime that are not often observable in the single-turn setups?
    \item How can LLM Agents be taught to more efficiently interface with interactive tools?
\end{enumerate}



Through our work, we demonstrate that state-of-the-art models like GPT-5 achieve only 66\% accuracy on the text-only split of \benchabbr~. This performance drastically decreases to approximately 36\% when all modalities are taken into account simultaneously. This highlights that \textbf{LLMs are poor at asking high-quality follow-up questions} under our interactive setup; specifically, a common failure mode is querying the Memory Agent for additional information in vain, therefore derailing the reasoning process. However, our ablations show that \textbf{models can still improve their performance with the assistance of the Memory Agent}, and we discuss a potential approach for improving model performance on \benchabbr. 

Overall, our work proposes a \textit{novel, multi-turn, dual-agent evaluation setup} for tip-of-the-tongue search, aiming to measure clarification-seeking capabilities. Such interactive benchmarks offer more comprehensive evaluations, better preparing agentic systems for conversing with real-world users. We hope \benchabbr~will lay the foundation for the next generation of search agents, capable of reasoning about clarification questions and resolving real-world ambiguities.

\begin{figure*}[!h]
    \centering
    \includegraphics[width=0.7\textwidth]{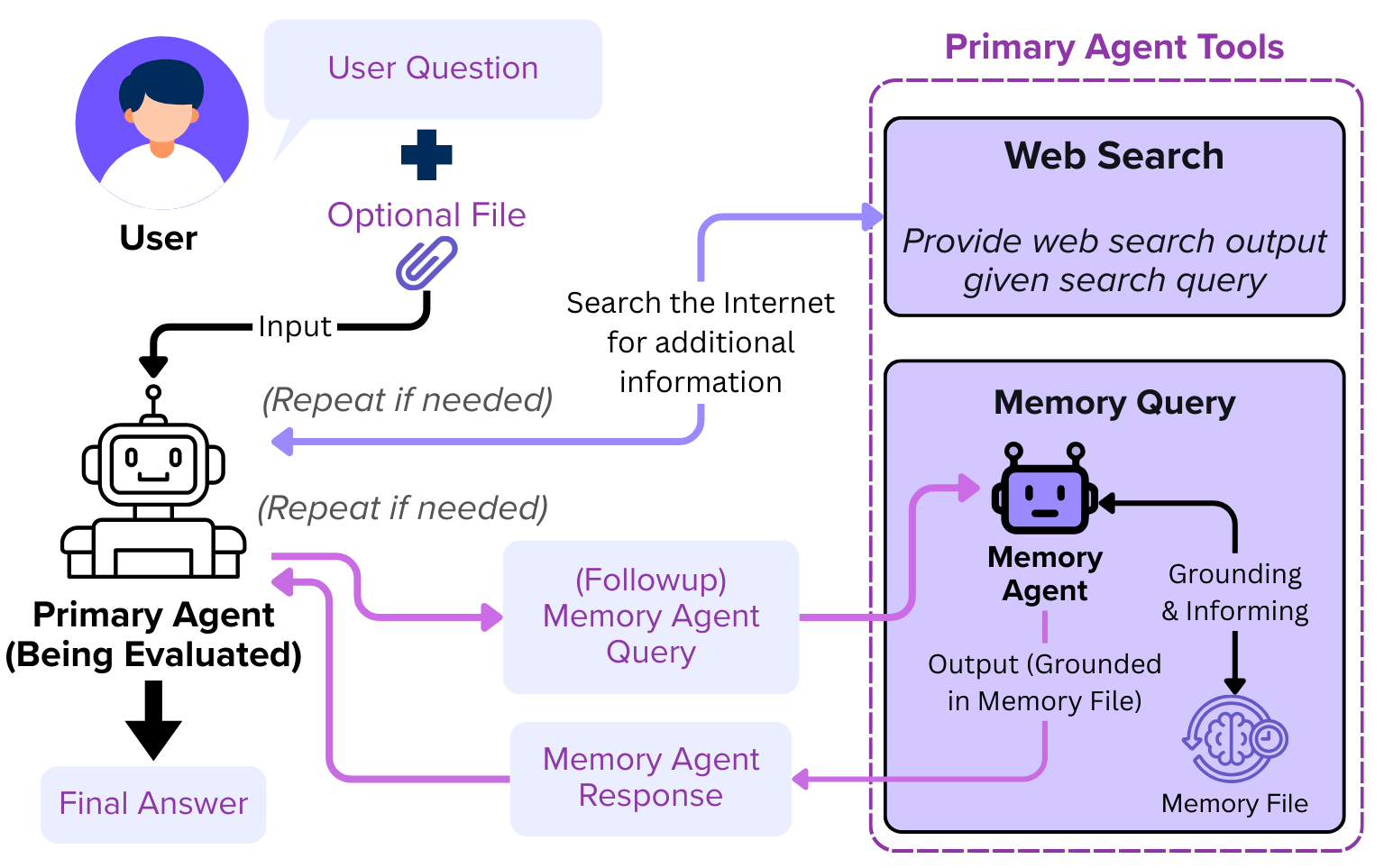}
    \caption{The overview of our dual-agent tip-of-the-tongue known-item search and retrieval process.}
    \label{fig:overview_process}
\end{figure*}

\section{Related Work}

\paragraph{Agentic Search Evaluation}
Web search benchmarks such as Mind2Web~\cite{gou2025mind2web} established evaluation for generalist web agents across diverse domains using real-world websites, while WebArena~\cite{zhou2023webarena} introduced a reproducible, self-hosted environment, emphasizing programmatic state checks over action sequence matching. However, as the popularity of deep research evaluations has grown~\cite{du2025deepresearch}, evaluations such as GAIA~\citep{mialon2023gaia} have introduced tasks that require tool use, multimodality handling, and web browsing, which remain simple for humans yet challenging for frontier models, relying on short output phrase match. Domain-specific benchmarks like \citet{he2025localsearchbench} focus on multi-hop local service search, shopping~\cite{lyu2025deepshop}, medicine~\cite{chen2025medbrowsecomp} or finance~\cite{li2024agent}. While these benchmarks reveal important trends in the deep research space, they often fail to account for characteristics such as ambiguity, which are common in real-life queries.  

\paragraph{Searching Under Uncertainty}

The importance of uncertainty in known-item retrieval queries has been more recently recognized~\cite{he2025tip, sky2025browsing}. \citet{sky2025browsing} specifically shows that ambiguity resolution in real-world search queries is challenging for state-of-the-art web agents. Soon after,~\citet{deng2025interactcomp} found that LLMs frequently fail to recognize ambiguity in search queries due to systematic overconfidence, and that open-source models are aligned to minimize response uncertainty rather than seeking clarification. 
However, this work does not study the quality of agent follow-up questions or their effectiveness in disambiguation. \benchabbr, on the other hand, encourages agents to ask relevant follow-up questions to further study an agent's ability to clarify tip-of-the-tongue queries in a multimodal setup.     

\paragraph{Interactive Benchmarking of Agents} Conversational search remains a key unsolved problem for modern AI systems~\cite{mo2025survey}. \citet{salle2022cosearcher} is among the first to propose a simulation environment to study clarification question characteristics for web queries. ~\citet{zamani2020analyzing} shows that asking good-quality clarification questions can improve search performance. 
The MIMICS dataset~\cite{zamani2020mimics} provides clarifying questions with human quality labels, paired candidate answers, and associated landing pages, while \citet{tang2025clarifying} proposes an ambiguity-driven chain-of-thought approach for meaningful disambiguation. However, these benchmarks are largely static; grounded evaluation of user preferences requires dynamic simulation~\cite{balog2025user}.
In this spirit,~\citet{barres2025tau2} designs a multi-turn simulation environment to deterministically evaluate the tool using LLM agents. Building on these principles, \benchabbr~aims to simulate a \textit{Memory Agent} to more realistically represent and ground user query clarifications.  

\section{\benchabbr}

\benchabbr~is a benchmark for evaluating multi-turn agentic search capabilities. To simulate a realistic evaluation scenario, we model tip-of-the-tongue search in an interactive setting, where one speaker is attempting to recall a specific entity. The other speaker would prompt the person performing the recollection with various questions to guide them to the target entity. 

We operationalize the process of recalling a specific entity -- tip-of-the-tongue known-item retrieval \cite{arguello2021tip} -- by modeling a dual-agent framework consisting of a \textit{Primary} and a \textit{Memory Agent}. The Primary Agent is equipped with a memory tool for querying the Memory Agent with free-text questions, as well as a web search tool. For each \benchabbr question, the Primary Agent has access to an optional input file. The Memory Agent receives a memory file containing often abstract cues about the target entity; the memory file does not directly refer to the answer to prevent information leakage. When responding to queries, the Memory Agent is instructed to ground all responses in the memory file only. Both the web search tool and the memory tool can be called repeatedly to provide clarifying information. See an overview of this process in Figure \ref{fig:overview_process}.

\begin{figure*}[!ht]
    \centering
    \includegraphics[width=0.8\linewidth]{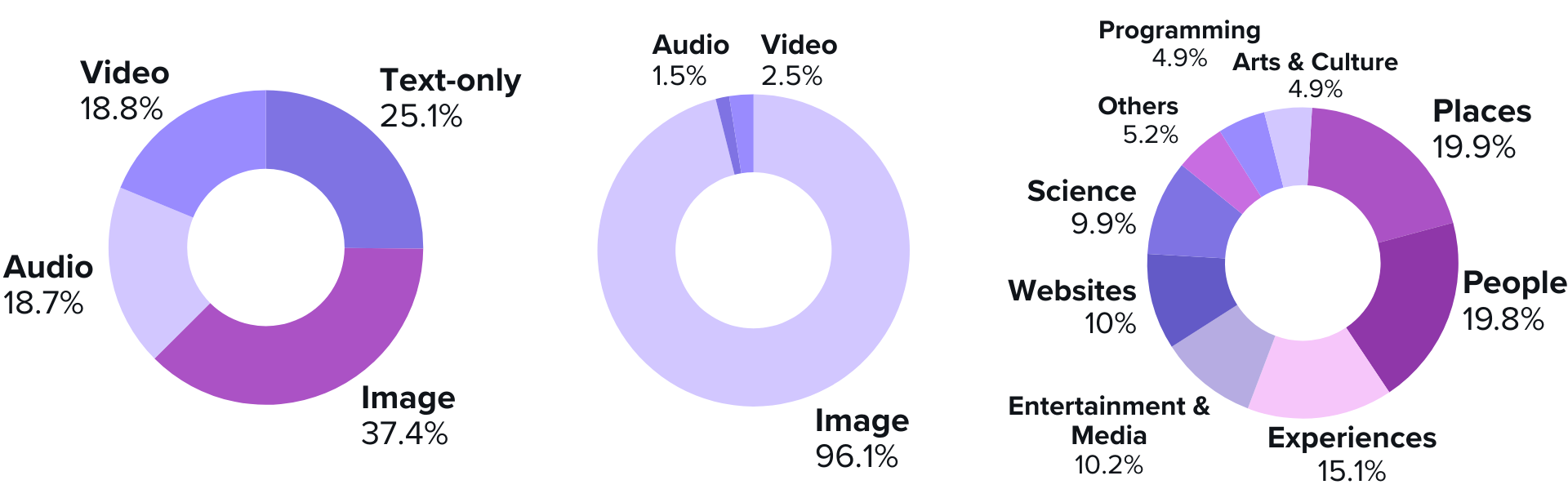}
    \caption{Distributions of input modality (left), memory file modality (middle), and question domains (right).}
    \label{fig:dist_mod}
\end{figure*}

\subsection{Data Collection}

Professional annotators were hired from an AI data curation company at \$20 per hour. We instructed the annotators to remember a time when they could clearly recall the sight or sound of an entity, yet failed to pinpoint exactly what it was called. The annotators are responsible for \textit{formulating the queries}, \textit{collecting} corresponding \textit{input and memory files}, and \textit{creating thought chains} that simulate the Primary Agent's process of solving the queries using web search and memory query tools. We define the following desiderata for \benchabbr:
\begin{enumerate}
    \item \textbf{Answer Non-Ambiguity:} Answers to the queries should be unambiguous. Each question in \benchabbr~is \textit{validated by a second annotator} to ensure that the final solution is consistent with the original answer.
    \item \textbf{Prompt Uncertainty:} To realistically simulate the tip-of-the-tongue search scenario, the questions supplied should be underspecified and incomplete.
    \item \textbf{Reasonable Complexity:} Queries should require extensive reasoning and synthesizing information intelligently from multiple sources. Annotators have been instructed to ensure that their questions cannot be answered through one internet search. To make the quality assurance process more manageable, the amount of time required for a human to answer each query is kept under 30 minutes.
    \item \textbf{Answer Longevity:} Answers should remain the same over time. Solving the questions should not require subjectivity or knowledge that may evolve with time.
\end{enumerate}

To comply with copyright restrictions, annotators may only collect input and memory files from public-domain sources, such as Pixabay\footnote{\url{https://pixabay.com/}}. When a question involves recalling a copyrighted song, annotators typically hum the melody themselves rather than copying from recordings. They are also instructed not to disclose any personally identifiable information to minimize the privacy risk associated with our data.

Additional measures were employed to increase the difficulty of \benchabbr. Annotators can optionally mix in \textbf{intentional errors} in their queries, consistent with BLUR: for instance, referring to a famous battle as occurring in Europe, while it occurred in the United States.  Furthermore, we supply the annotators with a \textbf{\texttt{\texttt{GPT-5}}-powered sanity check tool}: given the query and the file input, we prompt \texttt{GPT-5} using the tip-of-the-tongue search agent prompt from \citet{sky2025browsing} to see if \texttt{GPT-5} can successfully arrive at the correct answer utilizing OpenAI's web search tool. Annotators are encouraged to revise their prompts if \texttt{GPT-5} correctly answered without Memory Agent assistance. The authors and the annotators held weekly meetings to update any requirements and discuss any confusion throughout the dataset creation process.

\subsection{Benchmark Composition}

The final version of \benchabbr~contains 1,011 questions spanning eight domains. Of the input files, 254 were text-only, 378 contains image file inputs, 189 audio files, and 190 videos. As for the memory files, most of them are images (96.0\%), with audio and video files occupying 1.50\% and 2.50\%, respectively (Figure \ref{fig:dist_mod}). We additionally chart \benchabbr~composition by domain in Figure \ref{fig:dist_mod}.

\section{Evaluation Harness}

Due to the interactive nature of \benchabbr, the prompt of the Primary Agent and the implementation of the Memory Agent must stay consistent when evaluating different models. For the Primary Agent prompt, we modify the system prompt from \citet{sky2025browsing} to include instructions to leverage the memory agent, as well as an in-context example to facilitate advanced, abstractive reasoning. We detail this prompt in Appendix \ref{app:primary_agent}, and perform an ablation experiment on the content of the in-context example in Section \ref{subsec:ablation_examples}.

\subsection{Evaluating Primary Agent Correctness}
Similar to BLUR, we utilize LLM-as-a-judge to evaluate the correctness of the Primary Agent. Specifically, \texttt{GPT-4o} is used for its balance of cost-effectiveness and performance. The judge determines whether the Primary Agent's answer, often in the form of a series of reasoning steps arriving at a final result, is the same as the ground truth answer created by the annotator. We leverage the following prompt:

\lstset{
    basicstyle=\ttfamily\small,   
    breaklines=true,             
    breakatwhitespace=true,
    columns=fullflexible,
    keepspaces=true,             
    showstringspaces=false,
}
\begin{lstlisting}
Given a ground-truth answer and an answer from a language model, determine whether the language model's answer is correct.
\end{lstlisting}

\subsection{Memory Agent Construction}
To ensure that \benchabbr~performance is reflective of the reasoning capabilities of the assessed models, we must confine all reasoning processes to the Primary Agent. Therefore, we restrict the Memory Agent to respond solely based on the content of the memory file, without performing any deductions. We fix the Memory Agent model to be \texttt{Gemini-2.5-Pro} for its multi-modal capabilities.

Preliminary experiments reveal that the Primary Agent may occasionally bypass this restriction, resulting in the Memory Agent partaking in reasoning. We resort to prompt optimization to mitigate this issue. In particular, we leverage DSPy's MIPROv2 optimizer \cite{opsahl2024optimizing} to propose a set of 20 alternative prompts for the memory agent. For each of these proposed prompts, we fix the Memory Agent's instructions to this prompt, and run the full evaluation harness on a held-out validation set of 150 \benchabbr~instances, collecting the queries to and corresponding responses from the Memory Agent. Finally, we employ \texttt{Gemini-2.5-Pro} to judge if each of the responses from the Memory Agent can be directly derived from the memory files. This process improves the percentage of correctly grounded Memory Agent responses from 83.1\% to 97.5\%. We include the selected Memory Agent prompt in Appendix \ref{app:memory_agent}.

\section{Experiments and Discussion}

We test the performance on \benchabbr~for models with tool-calling capabilities, including \texttt{Gemini-2.5-Pro} \cite{comanici2025gemini}, \texttt{\texttt{GPT-5}}, \texttt{Claude-sonnet-4.5}, \texttt{Llama-4-Scout} and \texttt{-Maverick} \cite{meta2025llama}, Qwen3-Next-80B-A3B-Thinking (\texttt{Qwen3}, \citet{yang2025qwen3}), and \texttt{Kimi-K2-Thinking} \cite{team2025kimi}. For the open-weight models, we utilize the Together AI API implementations\footnote{\url{https://api.together.ai/}}. Each model is supplied with a web search tool, and the temperatures are set to \texttt{0}. For \texttt{GPT-5}, we use the native \texttt{web-search-preview} tool, while leveraging the implementation of a DuckDuckGo-powered search tool\footnote{\url{https://ai.pydantic.dev/common-tools/\#duckduckgo-search-tool}} for the others. An additional set of results, where \texttt{GPT-5} uses the DuckDuckGo tool, is available in Appendix \ref{app:duckduckgo}. 

\subsection{Evaluation on Full Harness}
As models exhibit different multimodal processing capabilities, we report results separately for multimodal and text-only settings. When a model has been evaluated under the multimodal setting, we select the model's solutions to all text-only \benchabbr~instances to compute the text-only evaluation results for that model, instead of conducting new evaluation runs. Each model is evaluated three times on the evaluation setting with the highest number of possible modalities. We record the results in Table \ref{tab:multimodal_eval}.

\textbf{Most models perform poorly on \benchabbr, with the top accuracies hovering below 70\%.} Across board, we observe that model performance drops even further when adding image input files, indicating that the input files increase reasoning complexity. We additionally observe that while frontier closed-source models achieves the highest accuracies, \texttt{Kimi-K2-Thinking} beats \texttt{Gemini-2.5-Pro} in text-only evaluation.

\begin{table}[!ht]
    \centering
    \begin{tabular}{c|cc}
    \toprule
    \textbf{Model Name} & \textbf{Text}  & \textbf{Text + Image}\\
    \midrule
       Gemini-2.5-Pro & 42.13 $\pm$ 7.51 & 38.44 $\pm$ 4.40 \\
       \midrule
       GPT-5 & \textbf{66.85 $\pm$ 1.48} & \textbf{60.40 $\pm$ 0.94} \\
       Claude-4.5 & \underline{59.04 $\pm$ 5.69} & \underline{53.74 $\pm$ 3.28} \\
       Scout & 16.69 $\pm$ 3.25 & 12.35 $\pm$ 1.09   \\
       Maverick & 23.11 $\pm$ 3.46 & 18.73 $\pm$ 1.61  \\
       \midrule
       Qwen3 & 20.41 $\pm$ 4.02 & N/A \\
       Kimi & 45.62 $\pm$ 2.34 & N/A\\
    \bottomrule
    \end{tabular}
    \newline
    \vspace*{5pt}
    \newline
    \begin{tabular}{c|c}
    \toprule
       \textbf{Model Name}  & \textbf{All Modalities} \\
       \midrule
        Gemini-2.5-Pro & 36.17 $\pm$ 2.59 \\
        \bottomrule
    \end{tabular}
    \caption{Accuracy results with 95\% confidence intervals from evaluating on the full harness, including the memory agent. The means and standard deviations are established using three full evaluation runs.}
    \label{tab:multimodal_eval}
\end{table}

\begin{table}[!ht]
    \centering
    \begin{tabular}{c|cc}
    \toprule
    \textbf{Model Name} & \textbf{\# Query} & \textbf{Corr.}\\
    \midrule
    Human & 4.64 & N/A\\
    \midrule
       Gemini-2.5-Pro & 1.92 & -0.025 \\
       GPT-5 & 4.46 & -0.336***\\
       Claude-sonnet-4.5 & 8.86 & -0.431***\\
       Llama-4-Scout & 2.77 & 0.0006\\
       Llama-4-Maverick & 3.47 & -0.11***\\
       Qwen3 & 0.25 & 0.170***\\
       Kimi-K2-Thinking & 2.73 & -0.117**\\
    \bottomrule
    \end{tabular}
    \caption{The average numbers of times the model queries the Memory Agent, as well as the Spearman correlation between the count for a given query and model accuracy on that query. ***$p<0.001$, **$p<0.05$.}
    \label{tab:mem_tool_counts}
\end{table}


\begin{figure*}
    \centering
    \includegraphics[width=\textwidth]{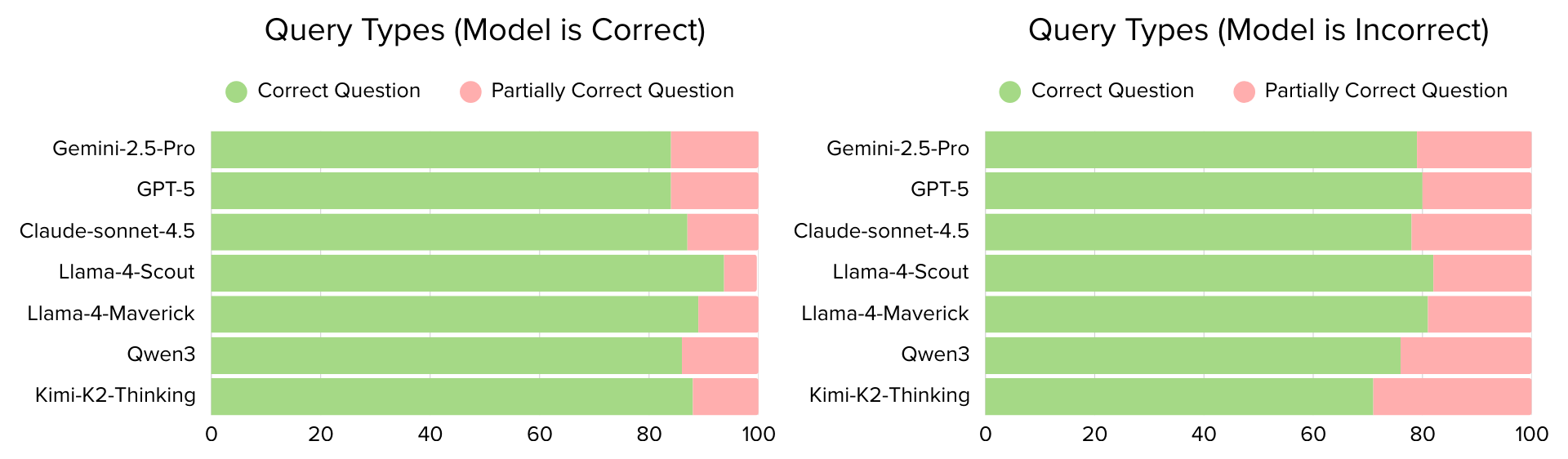}
    \caption{Query-type composition conditioned on model correctness. For each model, we show the proportion of questions that are fully correct (``Correct Question'') versus partially correct (``Partially Correct Question''), separately for cases where the model's final answer is correct (left) and incorrect (right). Bars are normalized within each model and correctness condition, so each row sums to 100\%.}
    \label{fig:err_types}
\end{figure*}

\subsection{Error Analysis}

We observe significant variance in attainable accuracies on \benchabbr. Inspecting the execution trajectories from the evaluation, we see that all models can struggle to leverage the memory agent to resolve ambiguities. We illustrate this with some example trajectories in Appendix \ref{app:ex_traj_mem}. In addition, we include a sequence of memory queries from \texttt{Claude-sonnet-4.5} on the \benchabbr~instance in Figure \ref{fig:ex_bench}, along with only a subset of the Memory Agent responses as the responses lean repetitive as the interaction continues:

\lstset{
    basicstyle=\ttfamily\small,   
    breaklines=true,             
    breakatwhitespace=true,
    columns=fullflexible,
    keepspaces=true,             
    showstringspaces=false,
}
\begin{lstlisting}
QUERY: What movie did I watch?
RESPONSE: I am unable to identify the movie you watched based on the provided file. The image does not contain any titles, text, or other direct identifiers that would confirm it is from a specific film. [Description of the memory file]
QUERY: Who are the actors in the movie?
RESPONSE: I am unable to provide an answer to your question. The provided file is a still image and does not contain any information to identify it as being from a movie, nor does it contain any data that would identify an actor. The face of the individual in the spacesuit is not visible. [Description of the memory file]
QUERY: What is the genre or theme of the movie?
QUERY: What is the title or name associated with this image?
QUERY: Is there any text, writing, or symbols visible in the image?
QUERY: Are there multiple figures or people in the image?
QUERY: What can you tell me about the celestial body or moon in the image?
QUERY: What movie title or name is shown in this file?
QUERY: What is the story or plot of the movie associated with this image?
QUERY: Is there any information about directors, producers, or studios associated with this image?
...
\end{lstlisting}

Here, \texttt{Claude} eventually incorrectly infers that the answer is \textit{Apollo 13} instead of \textit{The Martian}. We observe from the series of queries that, despite previous failed attempts, the Primary Agent model continues to probe the Memory Agent to extract direct information rather than performing more abstract reasoning. The Memory Agent often states that it is unable to answer the Primary Agent queries, but such refusals are ignored. This persistence can unnecessarily lengthen the interaction and confuse the Primary Agent, reducing its effectiveness in leveraging the Memory Agent. 

In order to quantitatively compare model behavior in terms of Memory Agent invocation, we compute the average number of times the memory query tool is called for each model (Table \ref{tab:mem_tool_counts}). For reference, we include the average number of hypothetical memory queries in the human-created thought chains. We additionally calculate the Spearman correlation between the count and whether the final response is correct (coded as 0 and 1).

Interestingly, there is often a \textbf{negative correlation between the number of memory queries and model accuracy}. Along with earlier examples of limited creative engagement with the Memory Agent, we believe this negative correlation arises because the Primary Agent fixates on extracting direct answers from it, repeatedly making unproductive requests. By contrast, \textbf{Qwen3, which rarely uses the Memory Agent, shows a positive correlation}, suggesting additional memory queries clarify rather than confuse in this case. Overall, our results demonstrate that current \textbf{models often fail to use the Memory Agent to disambiguate}, likely due to the abstract reasoning required.

Finally, as some \benchabbr~instances are injected with incorrect information, we want to determine whether this interplays with model performance. We compute the frequency of fully versus partially correct questions when each model arrives at the correct versus incorrect final answer. The results are documented in Figure \ref{fig:err_types}.

Generally, we observe that the percentages of incorrect \benchabbr~questions are higher when the model fails to answer correctly. This is expected, as models may become fixated on incorrect information, which can derail the reasoning process. However, for all models, we still observe non-zero proportions of correctly answered questions being partially correct. This indicates that models are still capable of resolving the injected errors through deliberation, although the model's ability to overcome injected erroneous information may vary.

Our conclusion is that \textbf{while models are more prone to incorrect answers with misleading information present, they \textit{can} sometimes successfully resolve these errors}. 

\begin{table}[!ht]
    \centering
    \begin{tabular}{c|cc}
    \toprule
    \textbf{Model Name} & \textbf{Text}  & \textbf{Text + Image}\\
    \midrule
       Gemini-2.5-Pro & 19.45 & 19.38 \\
       \midrule
       GPT-5 & \textbf{31.20} & \textbf{28.94}\\
       Claude-sonnet-4.5 & \underline{24.44} & \underline{22.81}\\
       Llama-4-Scout & 8.48 & 7.31\\
       Llama-4-Maverick & 10.76 & 9.47\\
       \midrule
       Qwen3 & 18.80 & N/A\\
       Kimi-K2-Thinking & 26.57 & N/A\\
    \bottomrule
    \end{tabular}
    \newline
    \vspace*{5pt}
    \newline
    \begin{tabular}{c|c}
    \toprule
       \textbf{Model Name}  & \textbf{All Modalities} \\
       \midrule
        Gemini-2.5-Pro & 19.44 \\
        \bottomrule
    \end{tabular}
    \caption{Accuracy results from evaluating \textbf{without the Memory Agent}. Each evaluation is run only once.}
    \label{tab:multimodal_nomem}
\end{table}

\subsection{Ablation: Memory Agent Query Tool}
We are interested in examining whether the Memory Agent is, in fact, critical to achieving better performance on \benchabbr. Therefore, we utilize the same prompt defined in \citet{sky2025browsing} to examine the capability of models to pinpoint the correct answer without Memory Agent assistance. We conduct this evaluation on the same set of models and include the results in Table \ref{tab:multimodal_nomem}.

\begin{figure*}[!ht]
    \centering
    \includegraphics[width=\textwidth]{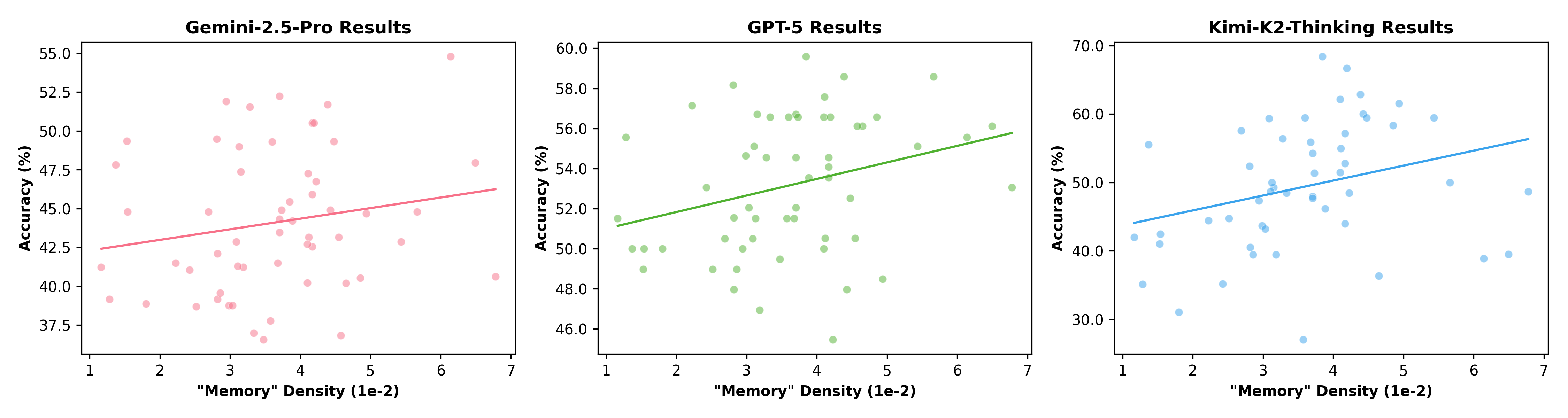}
    \caption{The results from our in-context example ablation experiment.}
    \label{fig:in_context_ex}
\end{figure*}

Comparing the accuracies between the evaluation runs, \textbf{performance is significantly worse without the Memory Agent}, highlighting its necessity in our evaluation harness and the difficulty of our task. However, models suffer varying extents of performance damage. This further illustrates that models may differ in reliance on the Memory Agent, as discussed in the error analysis section. Specifically, we observe that \texttt{Qwen3}, the model with the lowest average number of memory queries per question, suffers the least amount of accuracy reduction (1.61\%), but also has one of the lower accuracies.

The fact that it is still possible to deduce the correct answer without Memory Agent disambiguation (none of the models reach 0\% accuracy) indicates that \benchabbr~difficulty can be increased. We generally observe that answerable queries without a Memory Agent often correspond to the most ``popular'' options. See Appendix \ref{app:ex_traj_guess} for an example where the Primary Agent answers correctly through frequency-based assumption. To further increase the difficulty, we could enhance the ambiguity present in the questions by removing even more information from both the textual questions and the input files, which we leave to future work.

\subsection{Ablation: In-Context Examples}
\label{subsec:ablation_examples}

As observed in previous experiments, all evaluated models struggle to effectively interface with the Memory Agent, possibly due to not having been exposed to interactive tools during training. Because in-context examples are used to highlight desirable behaviors \cite{brown2020language,dong2024survey}, we aim to investigate whether using in-context examples that reason about the memory agent more extensively can improve performance.

To better assess the effect of the example, we randomly select 56 text-only question candidates and collect their annotators' thought traces. Then, we set the example question and in-context example thought trace in the Primary Agent system prompt to be each of these candidates, and evaluate each resulting agent on the last 100 modality-compatible \benchabbr~instances. This evaluation set does not include the question candidate. We perform this test for \texttt{Gemini-2.5-Pro}, \texttt{\texttt{GPT-5}}, and \texttt{Kimi-K2-Thinking}. This process yields 56 datapoints per model, as we compute the average accuracies for each in-context example candidate. Note that we focus on comparing \textit{within} models rather than \textit{between} models here, as models have different modality capabilities. We detail the variations in LLM-judged accuracies due to changes in in-context examples in Figure \ref{fig:in_context_ex}.

\textbf{Varying the in-context examples can significantly impact the resulting accuracy}. \texttt{Gemini-2.5-Pro}'s accuracy in this experiment ranges from 36.6\% to 54.8\%, a staggering 18.2\% difference. Similarly, \texttt{GPT-5}'s accuracy fluctuates between 45.5\% and 59.6\%. Variations are even more drastic for \texttt{Kimi-K2-Thinking} (\textbf{27.0\% to 68.4\%}). This demonstrates the key role the in-context example plays in our evaluation setup. 

Models are likely capable of extracting useful information from the demonstrations; we are now interested in what makes an example useful when evaluating on \benchabbr. We use a simple proxy for how much the in-context example is reasoning about the Memory Agent: the number of times the word ``memory'' has appeared in the thought trace, divided by the number of words in the thought trace. Concretely, we define:
\begin{align*}
    R_{\text{mem}} = \frac{\sum_{i}^N \mathbbm{1} \{w_i = \textbf{``memory''}\}}{N}
\end{align*}

where $N$ is the total number of words in the thought trace. The numerator would include mentions of the word ``memory'' from both explicit invocations of the memory query tool and the Primary Agent's deliberation when processing the Memory Agent's responses.

Finally, we compute (1) Spearman's correlation between $R_{\text{mem}}$ and the Primary Agent accuracy, and (2) the ordinary least squares regression results, modeling accuracy as a function of $R_{\text{mem}}$. We document this in Table \ref{tab:in_context_coeff}. 
Across the board, we see \textbf{positive correlations between the density of discussion about the memory agent in the in-context examples ($R_{\text{mem}}$) and the accuracy}. Though not always statistically significant, this suggests models benefit from extra memory query tool usage demonstrations, likely extending to other interactive-tool benchmarks.

\begin{table}[!ht]
    \centering
    \begin{tabular}{c|cc}
    \toprule
    \textbf{Model Name} & \textbf{Spearman's}  & \textbf{OLS}\\
    \midrule
       Gemini-2.5-Pro & 0.207 & 68.22 \\
       GPT-5 & 0.290* & 82.55* \\
       Kimi-K2-Thinking & 0.250 & 111.00 \\
    \bottomrule
    \end{tabular}
    \caption{The Spearman's correlation $\rho$ values and the coefficients from Ordinary Least Squares between $R_{\text
    {mem}}$ and average accuracy. * $p < 0.05$.}
    \label{tab:in_context_coeff}
\end{table}

Overall, we demonstrate that in-context examples offer a promising initial step toward enhancing model performance on \benchabbr. That said, the gains in accuracy remain relatively modest, suggesting that substantial further research is still required to teach models to reliably leverage agent tools with highly interactive components.

\section{Conclusion}


In this work, we propose \benchabbr, a multi-turn, interactive benchmark for tip-of-the-tongue search. We enlist human annotators to construct the queries and validate the answers, and introduce a dual-agent evaluation protocol where a \textit{Primary Agent} uses web search and a \textit{Memory Agent} to resolve ambiguous recollections. 

Our experiments illustrate that current models continue to struggle in this setting, with notable performance drops in multimodal regimes and frequent failures resulting from ineffective follow-up questions and unproductive memory querying. We find that in-context examples that engage with the Memory Agent more could enhance model performance on our task.

In conclusion, our work serves as an initial exploration into interactive benchmarking for tip-of-the-tongue search. We encourage the development of more interactive benchmarks that better capture real-world conversational search behaviors, and hope \benchabbr~can provide a standardized testbed for improving clarification strategies, reasoning, and tool use in future agentic search systems.

\section*{Limitation}
Even though our benchmark is sufficiently challenging for state-of-the-art models, instances in \benchabbr~can still be answerable even without Memory Agent assistance. We attempted to mitigate this during the data collection process by providing the annotators with the sanity check tool and encouraging them to modify their questions until \texttt{GPT-5} sans the Memory Agent fails the answer correctly. As previously mentioned, we leave increasing the difficulty of \benchabbr~even further as future work. 

While we performed some initial analysis into what makes in-context examples particularly useful to models under the \benchabbr~evaluation settings, our examination remains at the surface level. We only looked into a very simple word-based metric, namely the density of the word ``memory'' being mentioned. This measure does not necessarily reflect the actual quality of the in-context examples, i.e., whether the logical transitions between intermediate steps in the thought traces are valid and natural. Although examples that are high in our proxy metric may teach models how to interact with the Memory Agent in a human-like manner, they may still exhibit incorrect logic and lead models down rabbit holes. Future analysis work studying the relationship between the in-context examples and model performance on \benchabbr~would focus on determining the \textit{actual reasoning quality} of in-context examples.

\section*{Ethical Considerations}

During the data collection process, we have attempted to minimize privacy risk by asking annotators to refrain from discussing personally identifiable content. However, a certain, albeit limited, amount of privacy risk remains. Specifically, annotators expose their voices when humming songs as contextual information for the queries, and speech qualifies as sensitive biometric data under GDPR. To mitigate this in the future, we would consider voice cloning technologies that transform the voices, preserving the core content of the audio clues while sufficiently masking speech characteristics.

To prevent usage of images, audios, and videos without author consent, \benchabbr~is built completely on license-free content. Unfortunately, the expressivity of materials in this category may be limited. This, in turn, restricts the achievable range of questions from \benchabbr, as some clues are too niche to represent with public domain material. 

\section*{Acknowledgements}
We are grateful to Annie Franco, Gregoire Mialon, Amine Benhalloum, Chris McConnell, and Rohit Patel for their useful and constructive feedback on the benchmark.

\bibliography{custom}

\appendix

\section{Primary Agent Prompts}
\label{app:primary_agent}
\subsection{With Memory Agent Query Tool}
We hold the following core instructions intact, and experiment with different in-context example traces in Section \ref{subsec:ablation_examples}.

\begin{lstlisting}
AGENT CONTENT: You are a general AI assistant assigned to help a user identify an entity they have forgotten. Below is the user's query, asked on August 23rd 2025, consisting of a textual description possibly accompanied by an optional file attachment. Approach the task methodically, utilizing all available tools, and document each step you take in detail. Conclude your response with the following format: Final Answer: [Your final answer]. Your final answer should accurately provide the answer to the user's query as a short string.

You are allowed to consult a memory agent using the `memory_query` tool. The memory agent, when prompted properly, will provide you with descriptions of a memory file. This will be helpful for you to identify the answer. The clues will not always be straight-forward, and you may need to use additional reasoning and association to solve the clue. You may need to utilize this tool multiple times before you can achieve a useful response.

Here is an example of how you may query the memory agent when answering a question:

ORIGINAL QUERY: <In-context example query>

PROCESS TO ANSWER THE ORIGINAL QUESTION:
<In-context example thought trace>

Now use a similar type of logic to answer the following query. Leverage the memory agent creatively and think abstractly.

Query:
\end{lstlisting}

\subsection{Without Memory Agent Query Tool}

\section{Best Memory Agent Prompt}
\label{app:memory_agent}

\lstset{
    basicstyle=\ttfamily\small,   
    breaklines=true,             
    breakatwhitespace=true,
    columns=fullflexible,
    keepspaces=true,             
    showstringspaces=false,
}
\begin{lstlisting}
Imagine you are a security agent responsible for providing strictly factual, file-based answers in a classified investigation. Your responses must be limited ONLY to what is directly observable in the provided file (such as an image, audio, or video), with absolutely NO inferences, guesses, or leak of sensitive or unrelated information. Each time you answer a question, consider that revealing extra details or speculating may lead to a breach of confidentiality. If you cannot provide an answer based strictly on the contents of the file, clearly state that you are unable to do so and explain what file-based information you can provide instead. Always ensure your answer is rooted solely in the evidence present in the file and does not draw upon outside knowledge or assumptions. The accuracy and security of this investigation depend on your adherence to these file-only, no-inference constraints.
\end{lstlisting}

\section{Example Trajectories}
\subsection{Primary Agent Struggling to Interact with Memory Agent}
\label{app:ex_traj_mem}

\subsubsection{Overview}
For this example, \texttt{GPT-5} fails to engage with the Memory Agent correctly, attempting to pass all of the reasoning directly to the Memory Agent.

\subsubsection{Original Query}
I’m trying to remember the name of a plugin for a popular website builder that was created in 2005. I attached the web builder’s logo. The plugin I’m trying to find allows you to build websites using a visual, drag-and-drop interface. It also had a fairly simple logo, I think. What’s the name of the plugin?

\subsubsection{Human-authored Reasoning Trace}

\begin{lstlisting}
1. Thought: Let me see what the website builder's logo looks like for more context.
2. File search: It looks like a logo with ""W"" boldly written, and surrounded by a circle.
3. Web search: Website builder with a W logo.
4. Thought: The result returned several web builders. However, only ""WordPress"" has a logo similar to the image in the file attached.
5. Web search: When was WordPress created?
6. Thought: The result confirms WordPress was created in 2003, not 2005 as stated in the prompt.
7. Web search: "WordPress plugin that allows you to build websites using a visual, drag-and-drop interface."
8. Thought: The result returned multiple plugins and page builders used in WordPress, with visible logos on some of them. The top three have logos with two depicting a leaf, while one is an animal.
9. Memory search: Does the logo have an illustration of an animal?
10. Memory response: I don't know, but the file contains an image of even sticks horizontally arranged.
11. Thought: Cross validates other logos in previous search for visual match. Found a plugin called ""Elementor Pro"" with a logo featuring even strokes that form the letter ""E".
12: Final Answer: Elementor.
\end{lstlisting}

\subsubsection{Sample Reasoning Trace from \texttt{GPT-5}}
\lstset{
    basicstyle=\ttfamily\small,   
    breaklines=true,             
    breakatwhitespace=true,
    columns=fullflexible,
    keepspaces=true,             
    showstringspaces=false,
}
\begin{lstlisting}
memory_query
MEMORY QUERY: We see a hand-drawn WordPress-style W in a circle. The user asks for a WordPress plugin created in 2005 that let you build websites with a visual drag-and-drop interface and had a simple logo. What memory file clue corresponds to this? Provide hints like keywords, year, notable features, or letters.
---
memory_query
MEMORY CONTENT: I am unable to provide an answer to your query. The provided file does not contain any information, keywords, years, notable features, or letters corresponding to a WordPress plugin, a logo, or the year 2005.

Based strictly on the provided image file, I can confirm the following observable details:

*   **File Contents:** The image displays at least three light-colored, rectangular wooden blocks.
*   **Arrangement:** The blocks are oriented horizontally and arranged in a vertical line.
*   **Background:** The background is a solid blue color.
---
memory_query
MEMORY QUERY: Think abstractly: For WordPress, what 2005 plugin introduced a visual drag-and-drop interface for arranging site elements? Hints: it later became a core feature by 2007; it focused on sidebars; it was called ______ Widgets. Provide the missing word or a clue.
---
memory_query
MEMORY CONTENT: I am unable to provide the missing word or a clue based on the provided file.

The file is an image and does not contain any text, data, or information related to WordPress, its plugins, or their development history.

Based strictly on the contents of the file, I can confirm the image contains multiple rectangular, light-colored wooden blocks arranged horizontally against a solid light blue background.
---
memory_query
MEMORY QUERY: Retrieve any memory clue for an entity created in 2005 with a simple logo that lets users build sites via drag-and-drop; associated with WordPress and a common UI term for small modules. Provide a keyword or riddle.
---
memory_query
MEMORY CONTENT: I am unable to retrieve the requested information based on the provided file.

The query asks for a memory clue, keyword, or riddle related to an entity from 2005, its function, its logo, and its association with WordPress and a UI term. This information is not present in the provided image file. To provide such a clue would require using external knowledge not contained within the evidence, which is a violation of protocol.

Based strictly on the provided file, I can report the following:

The file is a still image containing multiple rectangular, light-colored wooden blocks. The blocks are arranged horizontally, one above the other, against a solid blue background. No other details, text, or symbols are visible.
---
memory_query
MEMORY QUERY: Possible answer: 'Sidebar Widgets' (the early WordPress plugin for drag-and-drop widgets). Do any memory clues reference the word 'widget', 'sidebar', or the year 2005 to confirm?
---
memory_query
MEMORY CONTENT: Based on a strict analysis of the provided file, I am unable to provide an answer to your query.

The image file does not contain any text, words, or numbers. Therefore, I cannot confirm the presence of the words 'widget', 'sidebar', or the year '2005'.

What can be observed from the file is an image displaying multiple rectangular, light-colored wooden blocks arranged horizontally against a solid blue background.
---

AGENT CONTENT: PROCESS

1) Visual inspection of the attachment

- The hand-drawn "W" inside a circle strongly resembles the WordPress logomark, so the "popular website builder" is WordPress.

2) First hypothesis generation

- Early WordPress "visual, drag-and-drop" building was introduced through the Widgets system, originally shipped as a plugin that let users drag items into sidebars. That plugin was commonly called "Sidebar Widgets."

3) Memory agent checks (attempted multiple times)

- Query 1 to memory agent asking for any clue tied to a 2005 WordPress page-builder plugin: returned only an unrelated description of wooden blocks; no useful clue.

- Query 2 asking specifically for hints about "widgets/sidebar/2005": again returned only the same unrelated wooden-blocks description.

- Query 3 proposing "Sidebar Widgets" for confirmation: again no corroborating clue. Conclusion: memory agent provided no actionable hints for this task.

4) Web verification

- Automattic's old "How to use Widgets" page explicitly calls the plugin "Sidebar Widgets" and explains the drag-and-drop interface for arranging sidebar elements; it also notes that widgets were later included in WordPress 2.2, confirming it started as a plugin.

- A contemporary Wired write-up on the WordPress 2.2 release notes that widgets had previously been a plugin and were integrated into core in May 2007, aligning with the above.

- Blog posts from March 2006 discuss the release of the "WordPress Sidebar Widgets" plugin to the public, matching the name and function.

5) Synthesis

- The attachment points to WordPress.

- The plugin that matches "visual, drag-and-drop interface" and an early, simple identity is WordPress' original "Sidebar Widgets" plugin, which later became the core Widgets feature.

Final Answer: Sidebar Widgets
\end{lstlisting}

\subsection{Primary Agent ``Guessing'' Correctly without Memory Agent}
\label{app:ex_traj_guess}

\subsubsection{Original Query}

I watched a series back in 2020, even though it came out a couple of years earlier. The story was about a boy who was kept in a device to grow for an extended period. As a result, he became super smart, had some special powers. Please, help me find the name of this series.

\subsubsection{Human-Authored Reasoning Trace}

\lstset{
    basicstyle=\ttfamily\small,   
    breaklines=true,             
    breakatwhitespace=true,
    columns=fullflexible,
    keepspaces=true,             
    showstringspaces=false,
}
\begin{lstlisting}
1. Thought: What device did the boy grow up in? Let me check the attached image.
2. File search: Opened the file to see a baby in an incubator
3. Web search: Movie about a boy who grew up in an incubator for a very long time and had superpowers.
4. Thought: The result of this search is an extensive list of possible movies.
5. Memory query: Is the movie Chronicle or Doomsday?
6. Memory response: I'm not sure, but I can see a woman making border of each side of her navel, on her abdomen.
8. Web search: Movie about a boy who grew up in an incubator for a very long time and had superpowers, abdomen.
9. Thought: The first link is a clip from the series Kyle XY. Let me confirm that he was kept for a long time in an incubator and had super powers.
10. Web search: Kyle XY movie plot
11. Thought: This link confirms that he became super smart and it also had a picture of the lead actor; he had no belly-button. Let me search to confirm that he was born in an incubator
12. Web search: Kyle XY born in incubator
13: Thought: This page confirms that he was birthed artifically for a lengthy period of time.
14. Final answer: Kyle XY
\end{lstlisting}

\subsubsection{Sample Reasoning Trace from \texttt{GPT-5}}

\lstset{
    basicstyle=\ttfamily\small,   
    breaklines=true,             
    breakatwhitespace=true,
    columns=fullflexible,
    keepspaces=true,             
    showstringspaces=false,
}
\begin{lstlisting}
AGENT CONTENT: Here's how I approached your query:

* Key clues I extracted: you watched it in 2020; the show released earlier; the plot centers on a boy grown in a device/pod for a long time, resulting in super-intelligence and special powers.
* Candidates I considered and quickly checked:

  * Beyond (2017): boy wakes from a 12-year coma with powers, not grown in a device. Ruled out.
  * Titans (2018): Conner is a clone grown in a Cadmus containment pod, but he isn't notably "super smart," and the pod is more for cloning/containment than prolonged growth from embryo to teen. Not the best fit.
  * Kyle XY (2006-2009): boy was gestated for years in a lab tank/pod, emerges with extraordinary intelligence and abilities. Strong match.
* Verification:

  * Kyle XY's protagonist was created and "gestated in a tank" for about sixteen years by the Zzyzx project, leading to exceptional intellect and abilities (e.g., telekinesis, advanced cognition).
  * Series details: aired June 26, 2006-March 16, 2009, which explains why it felt "a couple of years earlier" when you watched in 2020-it actually premiered much earlier.


Final Answer: Kyle XY
---
\end{lstlisting}





\section{GPT-5 + DuckDuckGo Search Tool Results}
\label{app:duckduckgo}

We evaluated GPT-5 with the full harness, replacing the OpenAI-native \texttt{web-search-preview} tool with the DuckDuckGo search tool, as is the case with the other models. We find that this new evaluation setting yields 57.52\% accuracy on \benchabbr for text + image and 64.34\% for text-only. Both of these numbers fall outside of the 95\% confidence interval of GPT-5 evaluation results in Table \ref{tab:multimodal_eval}, when GPT-5 leveraged \texttt{web-search-preview} instead. Furthermore, both of these numbers are lower in comparison. This indicates that GPT-5 may be more effective in gathering needed information using its native tool compared to using a third-party search tool.

\end{document}